%% file: main.tex
\documentclass[11pt,a4paper]{article}
\usepackage[hyperref]{acl2021}
\usepackage{times}
\usepackage{latexsym}
\usepackage{comment}

\usepackage[utf8x]{inputenc}
\usepackage{coloremoji}
\usepackage{booktabs}
\usepackage{array}
\usepackage[edges]{forest}

\usepackage{microtype}

\aclfinalcopy %
\def\aclpaperid{1452} %

\title{Deciphering Implicit Hate: Evaluating Automated\\ Detection Algorithms for Multimodal Hate\thanks{~~This paper is accepted for publication in \emph{Findings of ACL}, 2021.}}

\author{Austin Botelho\thanks{~~Work performed while at the University of Oxford}\\
  Anti-Defamation League\\
  \texttt{abotelho@adl.org} \\\And
  Bertie Vidgen \\
  The Alan Turing Institute\\
  \texttt{bvidgen@turing.ac.uk} \\\AND
  Scott A.\ Hale \\
  University of Oxford and Meedan\\
  \texttt{scott.hale@oii.ox.ac.uk}\\}

\date{}

\begin{document}
\maketitle

\newcommand{\thing}{paper}

\begin{abstract}
Accurate detection and classification of online hate is a difficult task.
Implicit hate is particularly challenging as such content tends to have unusual syntax, polysemic words, and fewer markers of prejudice (e.g., slurs). This problem is heightened with multimodal content, such as memes (combinations of text and images), as they are often harder to decipher than unimodal content (e.g., text alone). 
This paper evaluates the role of semantic and multimodal context for detecting implicit and explicit hate.
We show that both text- and visual- enrichment improves model performance, with the multimodal model (0.771) outperforming other models' F1 scores (0.544, 0.737, and 0.754).
While the unimodal-text context-aware (transformer) model was the most accurate on the subtask of implicit hate detection, the multimodal model outperformed it overall because of a lower propensity towards false positives.
We find that all models perform better on content with full annotator agreement and that multimodal models are best at classifying the content where annotators disagree.
To conduct these investigations, we undertook high-quality annotation of a sample of 5,000 multimodal entries. Tweets were annotated for primary category, modality, and strategy. We make this corpus, along with the codebook, code, and final model, freely available. 
\end{abstract}

\section{Introduction}
Although its prevalence is fairy low \cite{Vidgen2019}, the effects of online hate can be deeply pernicious, risking real harm on targeted victims and their communities \cite{Muller2019, Guadagno2013}. %
A 2021 survey by Anti-Defamation League found that 81\% of Americans agree social media companies should do more to counter online hate \cite{Anti-DefamationLeague2021}.

Research into automated hate detection has primarily focused on explicit varieties. However, many purveyors of hate have adopted more complex and nuanced strategies, such as dog whistling: the use of intentionally ambiguous rhetorical techniques to express hateful messages which only some audiences will recognize.
For instance, calls by right-wing American political figures to ``protect the suburbs'' cloak racial grievances and concerns about whiteness in more prosaic concerns about community protection.  

Performance in online hate classification has improved substantially from static methods like GloVe and fastText through the use of context-aware word embeddings, in particular those computed by transformer-models with self-attention \cite{Badjatiya2019, Mozafari2020,Polignano2019, Sabat2019, Zampieri2019, Yang2019, Sohn2019, kennedy2020contextualizing, vidgen2020detecting}.
However, most hate detection models are text-only and cannot be applied to non-textual content (such as images and audio) and do not account for non-textual information contained in multimodal content (such as memes). 
This is a problem of both task definition and modeling; most hateful content training datasets do not take into account non-textual features when annotations are made, which means that non-textual systems cannot be trained and evaluated on them.
The lack of detailed- and expertly- annotated datasets means that many key aspects of multimodal content classification have not yet been explored.%

We address these gaps in research, making three primary contributions.
First, we present a newly annotated dataset of 5,000 multimodal tweets, with labels for primary category, modality, and strategy.   
We make the annotation guidelines, code, and best performing models publicly available.
Second, we show that as models better take into account contextualization, from context-invariant to context-aware and unimodal to multimodal, how accurately they detect hateful content significantly improves. Though the unimodal-text context-aware model performs the best on the implicit hate subtask, the multimodal model is better overall due to a lower propensity towards false positives.
Third, we show that all models perform considerably worse on ambiguous content (as determined by annotator disagreement).

\section{Related Work}

The networked structure of online platforms means that hate is often able to spread far beyond the author's original intended audience \cite{Walther2011}. These ``masspersonal'' \cite{OSullivan2018} networks blur the divide between public and private discourse, resulting in ``context collapse'' as multiple audiences converge towards a singular unbounded one \cite{Marwick2010, Boyd2017}. This can increase the social costs of spreading hateful messages as wider audiences and platform moderators may disapprove of this content. Consequentially, hateful actors are incentivized to employ implict rhetorical strategies to circumvent these costs. Whereas explicit forms of hate (e.g., slurs or calls to violence) are likely to draw attention, subtle forms of hate, such as dog whistles, can be effective in avoiding detection.

Dog whistles comprise a range of strategies anchored in polysemy including pseudo-factual claims \cite{Meddaugh2009}, normative statements \cite{Pettigrew1995}, coded hate terms \cite{Magu2017}, and artistic license and humor \cite{Milner2013} to create implied meanings. This enables hateful actors to target their messages at different audiences such that the hateful elements are only recognized by people who are predisposed to respond favorably \cite{Albertson2015}. This gives their speakers plausible deniability, allowing them to avoid any social, legal, or platform-based punishment for the content they produce.

Multimodal communication, in particular, memes, are susceptible to co-optation by hateful actors for use as dog whistles because of its ability to convey incongruent ideas through each modality (`modal dissonance'). Hateful content can be passed under the fa\c{c}ade of a shared, seemingly benign, meme macro (or `template') \cite{Zannettou2018}. For instance, \citet{Vidgen2019} describe how a non-hateful image (e.g., a group of Muslims in prayer) can be combined with a non-hateful text (e.g., the words `Woken up yet?') to express prejudice. If the words or images were changed to something benign then the meme would no longer be hateful). This has led to a culture of 'shit posting' and trolling \cite{Pelletier-Gagnon2018, Phillips2012}.

Previous research on automated detection of hate has primarily relied on unimodal approaches with text-based features. These features have been passed through a variety of classification models \cite{Fortuna2018, Schmidt2017}. Neural network architectures harnessing advances in convolutions \cite{Gamback2017, Ribeiro2019, Zhang2018}, recurrence \cite{Pitsilis2018}, long-term memory \cite{Badjatiya2019, Pitsilis2018}, and bidirectionality \cite{Caselli2018, Qian2018} have been applied to improved accuracy. However, a shift towards fine-tuning large, pre-trained models has yielded the best results with BERT and its varieties being the models of choice \cite{Sohn2019, Zampieri2019, Mozafari2020, Polignano2019, vidgen2020detecting, vidgen-etal-2021-introducing}. 

Despite these strides, many challenges persist as real-world interactions are noisy, varied, and multimodal. Most applications of multimodal hate speech detection have combined text with meta-information like user characteristics, comment thread information, and network connections \cite{Chandrasekharan2017, FehnUnsvag2018, Gao2017, Qian2018, Vijayaraghavan2019}. Early examples of combining text and image data yielded mixed results \cite{Gomez2020, Sabat2019, Yang2019} leading companies like Facebook to initiate financial awards for improved performance \cite{Kiela2020}. Success in other domains like identifying pro-eating disorder content \cite{Chancellor2017}, gang activity on social media \cite{Blandfort2019}, demographic inference \cite{wang2019}, and cyberbullying \cite{Zhong2016} highlight the potential positive effects of multimodal approaches.

\section{Schema}
Our taxonomy comprises three main categories: (1) Primary attribute, (2) Modality, and (3) Strategy. The taxonomy and definitions were developed by reviewing existing theoretical frameworks for online hate and multimodal content \cite{Vidgen2019, waseem-hovy-2016-hateful, Kiela2020, Citron2011IntermediariesAH} and by iteratively investigating samples of tweets from the dataset. 

\paragraph{Primary}
For the \emph{Primary} attribute annotators selected one of four options: Hate, Counterspeech, Reclaimed, and Neutral. Similar to \citet[p.~512]{Davidson2017}, ``Hate'' is defined as ``language that is used to express hatred towards a targeted group or is intended to be derogatory, to humiliate, or to insult the members of the group.'' This definition's grounding in group identity differentiates hate from other forms of abusive content (such as interpersonal abuse) and corresponds with the definitions enforced by digital platforms like Facebook, Google, and Twitter \cite{YouTube2020, Twitter2020, Facebook2020}.

``Counterspeech'' is defined as any response to hateful speech that undermines it or expresses support to a group that it targets. This category is needed as models trained on datasets with only a `Neutral' category may struggle to differentiate between pro-social \cite{Galinsky2013} tweets and hateful ones if they have similar lexical content.
``Reclaimed'' is defined as the use of slurs self-referentially, whereby oppressive language is reappropriated for in-group use. This category is particularly important given well-established biases in classification models, whereby they disproportionately classify the vernacular of African Americans (and other groups) as hate \cite{Sap2019, Davidson2019}.
``Neutral'' is defined as content which did not fall into the other three categories.

\paragraph{Modality}
For the \emph{Modality} attribute annotators labeled the modality (image, text, or both) that was informative when making the \emph{Primary} annotation. This was needed because although all entries contained both a text and image, both modes were not always used to express hate. In some cases the hate was expressed solely by the text or by the image, and in others both were used together.

\paragraph{Strategy}
The \emph{Strategy} attribute captures the rhetorical devices used to convey hate. It expands upon the implicit/explicit distinction proposed by \citet{waseem-etal-2017-understanding} and adopted by others, such as \citet{casellietal2020} and \citet{Zampieri2019}.
Strategy is hierarchical: if annotators identify Hate then they can select ``Explicit'', ``Psuedo-factual'', ``Normative statements'', ``Coded language'', or ``Creative expressions'' (see Table \ref{tbl:examples}). The latter four were collapsed into a single ``Implicit'' category. These strategies reflect previous research on the varieties of implicit hate \cite{Meddaugh2009, Dvorak1999, Poynting2003, Hughey2013}.

\begin{table}
\begin{tabular}{m{.08\textwidth}m{.15\textwidth}m{.15\textwidth}}
\toprule
Strategy & Text & Image \\
\midrule
Explicit & \verb|<user>| \verb|<user>| suck a pig dick cunt \verb|<url>| & \includegraphics[scale=.15]{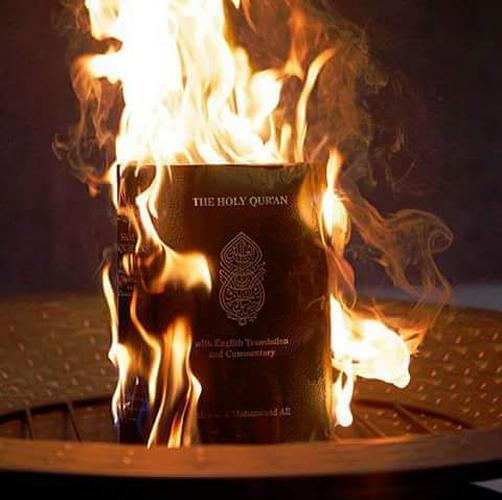}  \\
Pseudo-factual & \verb|<user>| illegal criminals protected by liberals. \verb|#buildthewall| \verb|<url>| & \includegraphics[scale=.15]{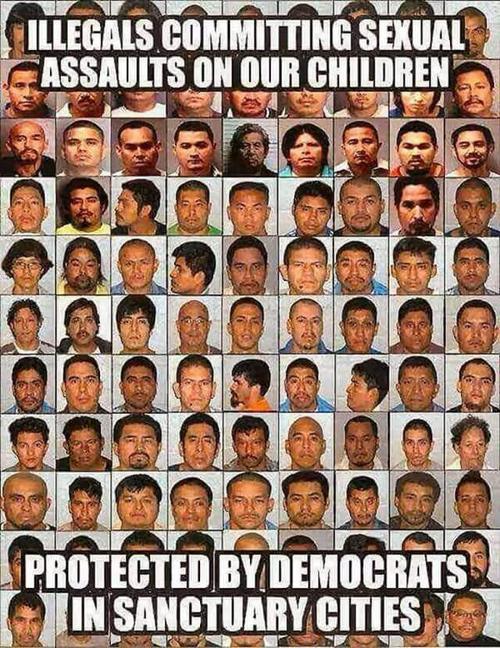} \\
Normative & \verb|<user>| \verb|<user>| come on booker bring in the ``race'' card you always do. \verb|<url>| & \includegraphics[scale=.12]{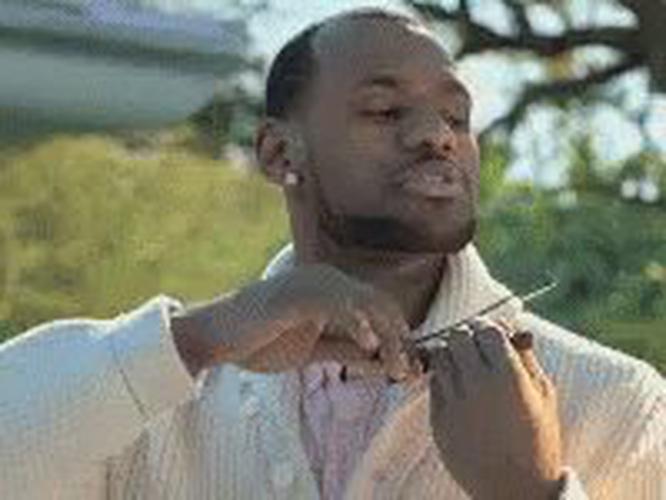}\\
Coded & best \verb|#npc| meme \verb|#npcmeme| \verb|#sjw| \verb|<url>| & \includegraphics[scale=.14]{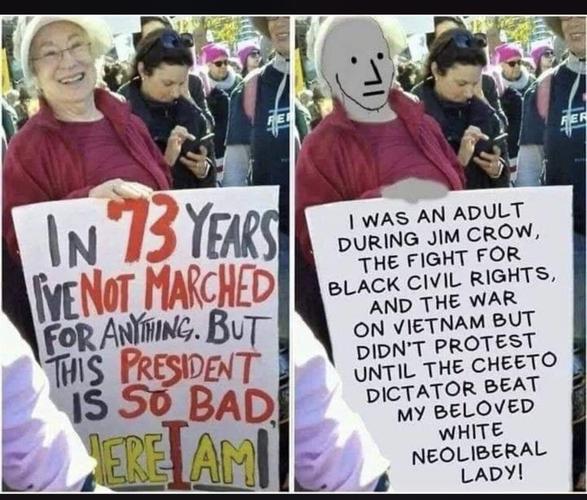}\\
Creative & good morning tweeps \verb|#friday| \verb|#teamtrump| \verb|#buildthewall| \redcircle \verb|⚪ | \bluecircle \verb|<url>| & \includegraphics[scale=.16]{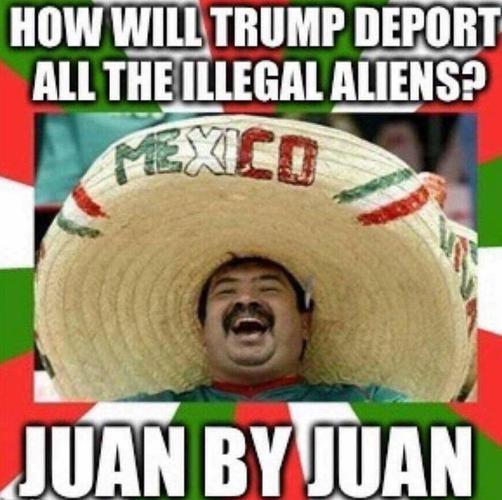}\\
\bottomrule
\end{tabular}
\caption{Examples of tweets for each implicit strategy.}\label{tbl:examples}
\end{table}

\section{Dataset and annotation}
The dataset used to train and evaluate the models originates from the MMHS150K collected by \citet{Gomez2020}. It comprises 150,000 English-language tweets which all contain text and an image, and have been annotated for hate (split into 5 subcategories: racist, sexist, homophobic, religion based attacks or attacks to other communities).\footnote{The dataset can be accessed at: \url{https://gombru.github.io/2019/10/09/MMHS/}}

We enrich a sample of 5,000 multimodal tweets from the MMHS150K dataset with re-annotation.\footnote{The 5,000 tweet dataset can be downloaded from \url{https://github.com/botelhoa/Dog_Whistle_Hate}}
Half of the dataset (2,500 tweets) was sampled using a 17-term query to provide more coverage of hateful tweets, especially covert ones.\footnote{The terms are: wall, card, confederate, maga, islam, sjw, gender, crim, npc, normie, ))), muslim, illegal, caravan, obama, hillary, america.} The other half (2,500 tweets) was randomly sampled to ensure heterogeneity and to offset biases associated with focused sampling \cite{WiegandRuppenhoferKleinbauer2019}.

\subsection*{Annotation process}
The annotators were given a detailed codebook to inform their decisions, with definitions, prototypical examples, and edge cases.
Each tweet was labeled by two annotators. All annotators had prior experience annotating online hate, and each completed a minimum of four weeks of training. Given the frequency of mislabeled hate speech due to a lack of domain expertise \cite{VanAken2018}, we prioritized annotator experience over more scalable crowdsourcing options like Amazon Mechanical Turk. All entries with disagreement were sent for review by an expert annotator. The expert was a PhD student researching online and offline hate, who had previously worked on two annotation projects.

The Kappa score for the dataset is 0.40, indicating low to moderate agreement. However, it is a nearly three-fold increase from the MMHS150k's Kappa of 0.15 despite the increased difficulty of the task. This is equivalent to other hate speech datasets. For instance, \citet{Wulczyn2017} report Krippendorf's alpha of 0.45 and \citet{sanguinetti-etal-2018-italian} report category-wise Kappas of 0.37 for \textit{offense} and 0.54 for \textit{hate}.

In 51\% of cases annotators disagreed on any of the \emph{Primary}, \emph{Modality}, or binarized \emph{Strategies} -- all of which were sent for review by the expert.
The annotators agreed on labels far more frequently for tweets in the ``None'' category (68.8\% of the time) than the others. Initial agreement (i.e. before expert adjudication) was 29\% for  Hateful, 15\% for Counterspeech and 14\% for Reclaimed. These relatively low agreement levels were primarily because broader situational context is often needed to make these judgments.

Annotators agreed less on ``Multimodal'' tweets than ``Unimodal-Text'' tweets. This suggests the richer semantic context from the different modalities helps annotators to clarify what is being expressed. Unexpectedly, agreement was higher when labeling implicit  rather than explicit hate (note that this is only for whether implicit hate was expressed, rather than identifying the particular technique used). This may be because the codebook contained more explanation of implicit hate, given we anticipated difficulties in annotating for them or because it can be difficult to ascertain when explicit slurs are hateful versus reclaimed when few signals pointing to the author's identity are available.

\begin{table}
\begin{center}
\begin{tabular}{l p{13em}}
\toprule
Annotations & Label Breakdown \\
\midrule
Primary & Hateful: 1850 (37.0\%),
Counterspeech: 113 (2.3\%),
Reclaimed: 366 (7.3\%),
None: 2,671 (53.4\%) \\
Modality &  Unimodal-text: 874 (37.5\%),
Unimodal-image: 25 (1.1\%),
Multimodal: 1,430 (61.4\%) \\
Strategy & Explicit:~31.8\%,
Normative:~30.4\%,
Coded:~22.3\%,
Creative:~7.8\%,
Psuedo-factual:~7.7\% \\
\bottomrule
\end{tabular}
\end{center}
\caption{Dataset label breakdown.} \label{tbl: label_breakdown}
\end{table}

\subsection*{Dataset composition}
The final labeled dataset is 37.0\% Hateful, 2.3\% Counterspeech, 7.3\% Reclaimed, and 53.4\% None (Table \ref{tbl: label_breakdown}).
Annotators relied on both modalities in a majority of cases (61.4\%).
In terms of strategy, of the tweets marked Hateful, 36.9\% were explicitly hateful and 63.1\% were implicitly hateful.
The implicit strategies were Normative claims (30.4\%), Coded language (22.3\%), Creative (7.8\%), and Psuedo-factual (7.7\%).  

\section{Model implementation}
All models were evaluated using the same 80/10/10 train, validation, and test split, stratified by class across the sets. Computation was completed using a single CUDA-enabled Nvidia Tesla K80 GPU in Google Colab.

\subsection{Input features}
The curators of the MMHS150k dataset represented all graphics (images, GIFs and video) as thumbnails. They were resized to be a pixel dimension of 500 in the smallest direction while maintaining the original aspect ratio. 
Textual features are derived from two sources: the tweet body and the image text, extracted using OCR. Both text sources underwent the same pre-processing procedure, using the Ekphrasis Python library.\footnote{The documentation is available at \url{https://github.com/cbaziotis/ekphrasis}} %
We de-noised the data by replacing hyperlinks, mentions, and dates with tags, decomposing hashtags into their constituent words, and normalizing elongated words and punctuation, in line with \citet{Mozafari2019} and \citet{Sohn2019}.
To retain indicators of sentiment, variables were added for whether capital letters were used, elongated words and punctuation \cite{Hutto2014}. The remaining text was truncated to a length of 100 tokens. Longer tweets were abridged while shorter ones were padded.

\subsection{Models}
Four classes of models with varying levels of semantic contextualization were trained, including three classes of unimodal models---unimodal-image models, context-invariant unimodal-text models (LSTM), and context-aware unimodal-text models (Transformers)---and multimodal models. Only the best performing model in each class is reported and their tuning described. Full results are available in the supplemental materials.

For unimodal-image models, Xception, NASNet, and Inception-ResNet V2 \cite{Chollet_2017, szegedy2016inceptionv4, Zoph_2018} were tested and Xception had the highest performance as measured by the weighted F1 score. For context-aware unimodal-text models, albert-xxlarge-v2, bert-large-uncased, electra-large-discriminator, and roberta-large \cite{lan2019albert, clark2020electra, Liu2019} were tested with roberta-large performing best. For multimodals, the intermediate concatenations approaches of \citet{Gomez2020} and \citet{Sabat2019} and a joint representation approach (``MMBT'') \cite{kiela2019supervised} were tested with the joint representation approach performing best.

\subsubsection{Unimodal-Image}
The first level of semantic contextualization consists of only image information. The extraction of image features was conducted with Xception \cite{Chollet_2017}. Xception decouples the mapping of cross-channel and spatial correlations by performing a depthwise convolution before a pointwise convolution. This improves Top-1 and Top-5 accuracy on ImageNet compared to Inception and a significant increase in performance on the larger JFT image corpus despite maintaining the same number of model parameters \cite{Chollet_2017}. It has yet to be applied to the task of hateful image recognition but outperforms methods that have \cite{Gomez2020, Yang2019, Sabat2019} in general image recognition tasks \cite{SooKo2020}.

The weights pre-trained on ImageNet were downloaded from the Keras library.\footnote{These weights can be found here: \url{https://keras.io/api/applications/}} Data augmentation was applied to the images in the train set prior to passing them through the network, including slight random rotations, height and width shifts, and horizontal flips. The images were passed in batch sizes of 32, following the approach of \citet{Szegedy2016}.
The weights in the bottom layers were frozen while updates occurred only on the top 5\%. A classifier was placed atop the CovNet which used two-dimensional Global Average Pooling followed by a fully connected layer of 1024 nodes with ReLU activation and a SoftMax output layer with dropout. During training, the same hyperparameters proposed by the original paper for the ImageNet task were applied with the addition of the early stopping regularization technique.

\subsubsection{Unimodal-Text}
In order to understand the impact of textual contextualization on the detection of implict hate, two approaches to text-only classification were implemented: (1) a context-invariant LSTM model and (2) context-aware transformer-based models. %

\paragraph{Context-Invariant LSTM}
The baseline context-invariant model consists of a bi-LSTM with pre-trained fastText embeddings \cite{Schuster1997}. This is the strongest alternative to a transformer model because it considers future context \cite{Graves2005} and approximately represents OOV words via character n-grams \cite{Bojanowski2016, Joulin2016}.

The LSTM comprised an embedding layer of length 300, two bidirectional LSTM layers of 256 hidden nodes with a dropout of 0.2, and a fully connected output layer. It was trained over 50 epochs with early stopping of a 10-epoch patience and 0.01 minimum average validation loss improvement using mini-batches of size 64 bucketed by length to reduce the need for padding. %
Parameters were optimized using the weighted ADAM algorithm \cite{loshchilov2017decoupled} with a Cross Entropy Loss function. A hyperparameter search was conducted across the learning rates $\in$ \{ 0.0001, 0.001, 0.01, 0.1, 1\}.

\paragraph{Context-Aware Transformer}
The context-aware model applies a transformer architecture, namely roberta-large~\cite{Liu2019}. Within the transformer, 
each use of a word is treated independently from its other uses. This means that, in principle, it can distinguish between the phrase ``race card'' when used in horse racing versus in reference to the view that racial prejudice can be advantageous to its victims.
The Transformer model was coded in PyTorch with pre-trained weights loaded through the HuggingFace library.\footnote{These are available at \url{https://huggingface.co/transformers/}.} %
It was trained for 10 epochs with early stopping of a two-epoch patience and 0.005 minimum average validation loss improvement. Parameters were optimized using the weighted ADAM algorithm \cite{loshchilov2017decoupled} with a 0.1 weight decay and slanted triangular schedule \cite{DBLP:journals/corr/abs-1801-06146} with a warmup of 0.06. Backpropagation was conducted using Cross Entropy Loss. A hyperparameter search borrowed from \cite{Liu2019} was implemented across the learning rates $\in$ \{ 1e-5, 2e-5, 3e-5\} and mini-batch sizes $\in$ \{ 16, 32, 64\}. 

\begin{table*}[t]
\begin{tabular}{l rrrr rrr rr}
\toprule
 & \multicolumn{4}{c}{Overall} & \multicolumn{3}{c}{Accuracy by Strategy}\\
 \cmidrule(r){2-5}\cmidrule(l){6-8}
& Accuracy & Precision & Recall & F1 & Non-Hateful & Explicit & Implicit\\
\midrule
Unimodal-Image & 0.604 & 0.560 & 0.544 & 0.544 & \textbf{0.760} & 0.273 & 0.370 \\
Unimodal-Text\\
~~Context-Invariant & 0.737 & 0.707 & 0.737 & 0.719 & 0.713 & 0.742 & 0.798 \\
~~Context-Aware & 0.765 & 0.759 & 0.765 & 0.754 & 0.678 & \textbf{0.864} & \textbf{0.941}\\
Multimodal & \textbf{0.785} & \textbf{0.763} & \textbf{0.785} & \textbf{0.771} & 0.732 & 0.833 & 0.899 \\
\midrule
Count & 502 & 502 & 502 & 502 & 317 & 66 & 119 \\
\bottomrule
\end{tabular}
\caption{Overall performance of models and performance split by strategy. The multimodal model performs best overall. The context-aware unimodal-text model performs best on both implicit and explicit forms of hate.} \label{tbl:models_performance_strategy}
\end{table*}

\subsubsection{Multimodal}
Lastly, deeper semantic contextualization may be achieved through the inclusion of multimodal data. Such models should, in theory, more accurately identify implicit hate by drawing from information contained by both the image, text, and their interaction. Recent approaches use a transformer's attention mechanism to generate joint representations of images and text \cite{kiela2019supervised, li2019visualbert, Lu_2020_CVPR}.

The MMBT first encodes image data using ResNet-152 with a generalized final pooling layer pre-trained on the ImageNet dataset. The image embeddings are combined with the tokenized text and passed through a bidirectional transformer architecture that was initialized using pre-trained BERT weights before an output layer with SoftMax activation makes the classification \cite{kiela2019supervised}. This was implemented using the Simple Transformers library.\footnote{https://github.com/ThilinaRajapakse/simpletransformers} A batch size of eight was used, with a learning rate of $1^{-5}$ trained with early stopping, weighted ADAM optimization \cite{loshchilov2017decoupled} with 0.1 weight decay, and a slanted triangular schedule \cite{DBLP:journals/corr/abs-1801-06146} with a warmup of 0.06.

\section{Results}

\subsection{Inter-modal Performance Comparisons}
Performance for the strongest models in each of the four modality types is displayed in Table \ref{tbl:models_performance_strategy}. Metrics were computed with a weighted average to accommodate class imbalances. The multimodal model performed the best across the four metrics, albeit only marginally so compared to the second best model, the unimodal-text context-aware model. Both unimodal-text models noticeably outperformed the unimodal-image model.

\subsection{Performance by Strategy}
To assess the secondary effects of defining training objectives, accuracy based on the \emph{Strategy} annotation was calculated. These labels were not shown to the model to demonstrate performance variations hidden by oversimplified label categories.

The unimodal-text context-aware model had the highest accuracy when identifying both ``Explicit'' and ``Implicit'' hate (Table \ref{tbl:models_performance_strategy}). By contrast, the unimodal-image model most accurately identifies non-hateful tweets, but struggles with hateful ones. 

\subsubsection{Ambiguity}
Ambiguity is the final characteristic for which performance was assessed (Table \ref{tbl:models_ambiguity}). A decision in the \emph{Primary} annotation is considered ambiguous if the two annotators provided conflicting decisions.

\begin{table}
\begin{center}
\begin{tabular}{lrr}
\toprule
\multicolumn{2}{r}{Unambiguous} & Ambiguous  \\
\midrule
Unimodal-Image & 0.758 & 0.398 \\
Unimodal-Text\\
~~Context-Invariant & 0.827 & 0.643  \\
~~Context-Aware & \textbf{0.875} & 0.682 \\
Multimodal & 0.845 & \textbf{0.692}   \\
\midrule
Count & 272 & 230 \\ 
\bottomrule
\end{tabular}
\end{center}
\caption{Model F1 Score by ambiguity. Entries where annotators disagree are considered ambiguous and entries with full agreement are considered unambiguous. All models perform better on Unambiguous content. The multimodal model performs best on ambiguous content.} \label{tbl:models_ambiguity}
\end{table}

All models had significantly higher F1 scores on data deemed unambiguous. The multimodal model dealt the best with ambiguity (0.692) followed closely behind by the unimodal-text context-aware model (0.682). The unimodal-image model was most affected (0.758 vs 0.398) by ambiguity.

\section{Error Analysis}
\begin{figure}[t]
\input{figs/tree-errors-forest}
\caption{Sources of classification errors for the multimodal model.
}
  \label{fig:error_analysis}
\end{figure}
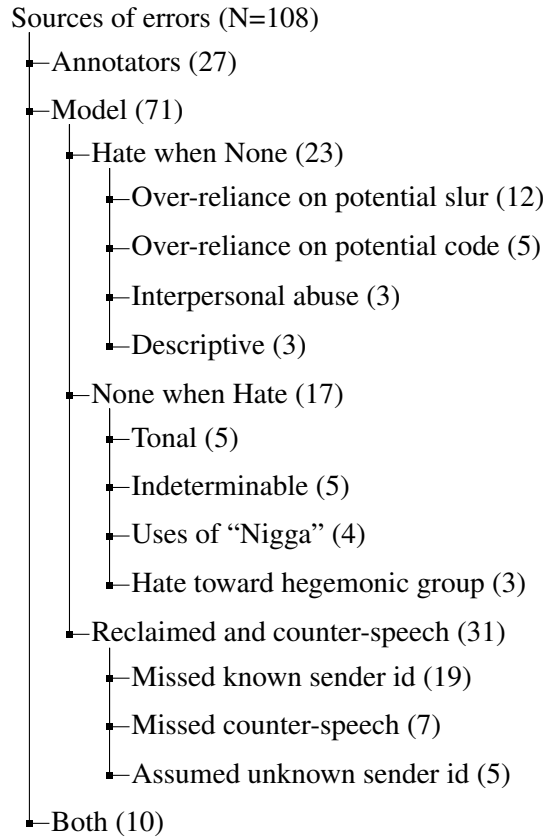

We conducted a qualitative analysis on the errors of the multimodal model similar to the one in \citet{vidgen2020detecting}. Errors were inductively categorized into ``mutually exclusive and collectively exhaustive'' groups \cite[p.~7]{vidgen2020detecting}. This is visualized in the Tree Diagram in Figure \ref{fig:error_analysis}.

The first branch on the tree is the \textit{annotator errors}. This occured when the model's classification better represented the tweet's content, based on the annotation codebook (as determined by this paper's authors). This represented 25\% of the errors and suggests that model performance could be improved further if annotator errors were eradicated.
Model errors were the most frequent, accounting for 66\%.
Lastly, in some cases we determined that the true label was a classification other than those provided by the annotators or the model, which account for 9\% of errors.

Model errors were further subdivided, as shown on the tree diagram branches.
\emph{Hate when None} includes instances where the model classified ``None" content as ``Hateful''. This was likely caused by an over reliance on the use of slurs (i.e., a slur being used non-hatefully), misidentification of interpersonal abuse or confusion caused by content which describes/reports on (but does not endorse) hateful activities.
\emph{None when Hate} is the inverse. It includes all instances where the model classified ``Hateful'' content as ``None''. This was likely caused by a combination of tone misjudgements (e.g., hateful language which had positive sentiment), uses of the word \emph{Nigga}\footnote{When the identity of the speaker was unknown and the use was not evidently ``Hateful'', annotators were instructed to treat the tweet as ``None'' to avoid penalizing language most commonly used by Black communicators. Ergo, the model may associate its use with ``None''.}, hate towards a hegemonic group (e.g., referring to White people as \emph{Hillbillies}), and, in select cases, indeterminable reasons. The final category, \emph{Reclaimed and Counterspeech}, comprises confusion beyond the ``None''--``Hateful'' distinction. These were caused by missing the sender's identity when it was knowable (i.e., ``Reclaimed'' language was classified as ``None''), assuming the sender's identity when it was unknowable (i.e., ``None'' content was classified as ``Reclaimed'', and failing to predict ``Counterspeech''. This final type did not exhibit systematic misclassifications, likely because there were few instances overall.

\section{Discussion}
As hypothesized, the unimodal-text context-aware model outperformed the unimodal-text context-invariant model. The 3.5 percentage point increase in F1 scores when distinguishing between \emph{Primary} labels (``Hate'', ``Counterspeech'', ``Reclaimed'', and ``None'') is in line with other results which range from 0--8 percentage point improvements depending on the dataset \cite{Vidgen2019, Sohn2019, Zampieri2019, Mozafari2019}.

When isolating performance by hate strategy, this gap substantially widens to 12.2 (``Explicit'') and 14.3 (``Implicit'') percentage points. This demonstrates the value of semantic contextualization for accurately identifying hate speech grows as its overtness diminishes. This is not the case with non-hateful content. On those, the unimodal-text context-invariant model is 3.5 percentage points more accurate.

Content modalities interact in ways that can create more subtle forms of hate. For this reason, the multimodal model was expected to outperform unimodal ones. This is marginally supported by the 1.7 percentage point gap in F1 score on the \emph{Primary} labels between the multimodal model and the best performing unimodal model. This improvement is larger than that achieved by \citet{Gomez2020, Sabat2019, Yang2019}, but smaller than that by \citet{Kiela2020} in similar hate speech detection tasks. Improvements from the incorporation of image data are minor compared to the gains from the transformer model. This may either signal underlying patterns in how multimodality is used in online content or reflect the data collection methods which relied on text-based query methods through the Twitter API.

However, it is the context-aware unimodal-text model that performs the best when assessing by \emph{Strategy}. The gaps between the unimodal-text model and multimodal model are 2.8 percentage points (``Non-Hateful''), 3.1 percentage points (``Explicit''), and 4.2 percentage points (``Implicit''). The models which only consider image data are more attuned to ``Non-Hateful'' tweets to the determinant of hate identification. This implies a dissonance between modalities which mitigates, sometimes rightly and other times wrongly, hateful signals from the text.

All models more accurately identify implicit than explicit hate. This is a surprising result which may reflect their higher representation within the training data rather than any inherent property that makes them more detectable.

\section{Conclusion}

This paper substantiates the need to consider varying forms of hate with different modalities and levels of overtness. It investigates the value of context-aware textual and multimodal features finding that both improve model F1 score with the multimodal model performing the best (0.771). Further, we find that model performance is directly contingent upon annotator agreement levels (referred to as `ambiguity' in the main body of the paper). These findings are generated from a newly-annotated dataset of 5,000 tweets containing information on each entry's primary attribute, modality, and strategy. This dataset along with the annotation codebook, model training code, and model weights are available to encourage future research on the topic.

\section*{Acknowledgments}
This work was supported by Wave 1 of The UKRI Strategic Priorities Fund under the
EPSRC Grant EP/T001569/1, particularly the ``Criminal Justice System'' theme within
that grant, and by The Alan Turing Institute. We are grateful to all our annotators and appreciate the feedback from our reviewers as well as that from the 2020 cohort of the MSc in Social Data Science at the University of Oxford.

\clearpage

\end{document}

%% file: figs/tree-errors-forest.tex
\newcommand{\eq}{=}
\begin{forest}
  for tree={
    grow'=0,
    child anchor=west,
    parent anchor=south,
    anchor=west,
    calign=first,
    edge path={
      \noexpand\path [draw, \forestoption{edge}]
      (!u.south west) +(7.5pt,0.5pt) |- node[fill,inner sep=1.25pt] {} (.child anchor)\forestoption{edge label};
    },
    before typesetting nodes={
      if n=1
        {insert before={[,phantom]}}
        {}
    },
    fit=band,
    before computing xy={l=15pt},
    inner sep=0.5pt,
  }
  [Sources of errors (N\eq108)
    [Annotators (27)]
    [Model (71)
      [Hate when None (23)
        [Over-reliance on potential slur (12)]
        [Over-reliance on potential code (5)]
        [Interpersonal abuse (3)]
        [Descriptive (3)]
      ]
      [None when Hate (17)
        [Tonal (5)]
        [Indeterminable (5)]
        [Uses of ``Nigga'' (4)]
        [Hate toward hegemonic group (3)]
      ]
      [Reclaimed and counter-speech (31)
        [Missed known sender id (19)]
        [Missed counter-speech (7)]
        [Assumed unknown sender id (5)]
      ]
    ]
    [Both (10)]
  ]
\end{forest}